\documentclass[10pt,twocolumn,letterpaper]{article}

\usepackage[utf8]{inputenc}
\usepackage{cvpr}              % To produce the CAMERA-READY version
\usepackage{graphicx}
\usepackage{amsmath}
\usepackage{amssymb}
\usepackage{booktabs}
\usepackage{xfrac}
\usepackage{subcaption}
\usepackage[accsupp]{axessibility} % Improves PDF readability for those with disabilities.

\usepackage[pagebackref,breaklinks,colorlinks]{hyperref}

\usepackage[capitalize]{cleveref}
\crefname{section}{Sec.}{Secs.}
\Crefname{section}{Section}{Sections}
\Crefname{table}{Table}{Tables}
\crefname{table}{Tab.}{Tabs.}

\def\cvprPaperID{16} % *** Enter the CVPR Paper ID here
\def\confName{CVPR}
\def\confYear{2022}

\begin{document}

%%%%%%%%% TITLE - PLEASE UPDATE
\title{Is Neuron Coverage Needed to Make Person Detection More Robust?}

\author{Svetlana Pavlitskaya, Şiyar Yıkmış and J. Marius Zöllner\\
FZI Research Center for Information Technology\\
76131 Karlsruhe, Germany\\
{\tt\small pavlitskaya@fzi.de, yikmis@fzi.de, zoellner@fzi.de}
% For a paper whose authors are all at the same institution,
% omit the following lines up until the closing ``}''.
% Additional authors and addresses can be added with ``\and'',
% just like the second author.
% To save space, use either the email address or home page, not both
%\and
%Şiyar Yıkmış\\
%FZI Research Center for Information Technology\\
%76131 Karlsruhe, Germany\\
%{\tt\small yikmis@fzi.de}
%\and
%J. Marius Z ̈ollner\\
%FZI Research Center for Information Technology\\
%76131 Karlsruhe, Germany\\
%{\tt\small zoellner@fzi.de}
}
\maketitle

%%%%%%%%% ABSTRACT
\begin{abstract}
The growing use of deep neural networks (DNNs) in safety- and security-critical areas like autonomous driving raises the need for their systematic testing. Coverage-guided testing (CGT) is an approach that applies mutation or fuzzing according to a predefined coverage metric to find inputs that cause misbehavior. With the introduction of a neuron coverage metric, CGT has also recently been applied to DNNs. In this work, we apply CGT to the task of person detection in crowded scenes. The proposed pipeline uses YOLOv3 for person detection and includes finding DNN bugs via sampling and mutation, and subsequent DNN retraining on the updated training set. To be a bug, we require a mutated image to cause a significant performance drop compared to a clean input. In accordance with the CGT, we also consider an additional requirement of increased coverage in the bug definition. In order to explore several types of robustness, our approach includes natural image transformations, corruptions, and adversarial examples generated with the Daedalus attack. The proposed framework has uncovered several thousand cases of incorrect DNN behavior. The relative change in mAP performance of the retrained models reached on average between 26.21\% and 64.24\% for different robustness types. However, we have found no evidence that the investigated coverage metrics can be advantageously used to improve robustness.
\end{abstract}

%%%%%%%%% BODY TEXT
\section{INTRODUCTION}

\begin{figure}
\centering
	\includegraphics[width=0.45\textwidth]{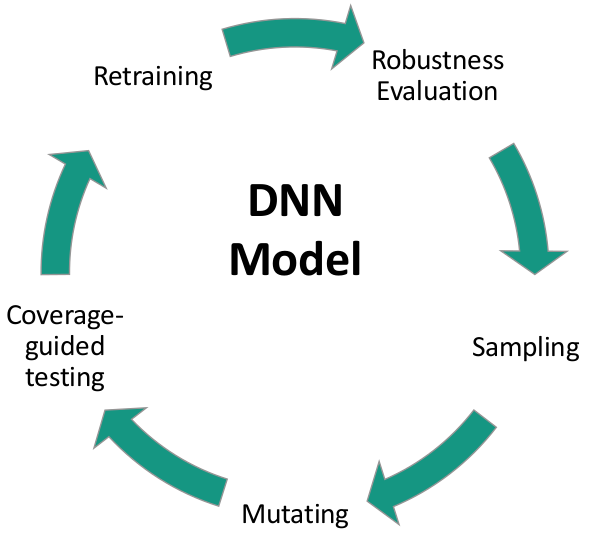}
	\caption{We propose repeated discovery of DNN bugs and subsequent model retraining to robustify DNNs for person detection against intended and unintended malfunctioning.}
	\label{concept_pic}
\end{figure}

Deep neural networks (DNNs) are increasingly used in safety- and security-critical areas. During operation, misbehavior of these systems can lead to fatal consequences for humans. Several test procedures such as fuzzing and coverage metrics from traditional software development have been adapted to the functionality of DNNs in order to identify bugs before deployment. Nevertheless, the suitability of coverage metrics for the detection of intended and unintended misbehavior of DNNs remains questionable.

This work aims to explore whether coverage-guided testing can adequately identify and fix DNN bugs. For this, we propose a pipeline, which includes finding DNN bugs and retraining the analyzed model on a training dataset, enhanced with found bugs. By adding coverage criteria to our definition of a DNN bug or excluding them, we can find out, whether coverage metrics correlate with DNN robustness.  Differently from previous work on coverage-guided DNN testing, we focus on person detection as a use case. The proposed pipeline helped to find several thousand cases of incorrect behavior. Retraining the model on the enhanced dataset helped to increase mAP by up to 64.24 percent points. We have however observed no robustness improvement when various types of neuron coverage were taken into account during bug finding. 

To the best of our knowledge, our work is the first to study the necessity of neuron coverage metrics in CGT for object detection models.

%While most existing work evaluates the correlation of neuron coverage with a better performance of image classification models, we focus on object detection task.

\section{RELATED WORK}

DNN testing is a method aiming to reveal faulty behavior like adversarial examples or to provide assurance cases~\cite{huang2020} by examining the model with a large set of test cases that are generated according to a predefined coverage metric and established techniques like input mutation or fuzzing~\cite{odena2019}. In order to apply coverage-guided testing, which is one of the established methods of software testing, to DNN systems, it is crucial to define a set of suitable coverage criteria. 

\subsection{Coverage Criteria}

One of the first metrics for the systematic testing of DNNs is the neuron coverage introduced by Pei et al.~\cite{pei2017}. It expresses the ratio of the number of unique activated neurons for all test inputs to the total number of neurons in the DNN. %Furthermore, Pei et al. formulate the problem of finding a large number of behavioral differences between similar DL systems while maximizing neuron coverage as a joint optimization problem which they solve with a gradient-based algorithm. Unlike the backpropagation algorithm, their gradient-based solution treats the input value as a parameter and the weight parameter as a constant.

Tian et al. \cite{tian2018} took up the idea of the neuron coverage metric and used it to generate tests for DNN-based autonomous driving cars. For this, domain-specific natural image transformations such as translation or blurring and metamorphic testing \cite{chen2020} were applied, which made the need for multiple DNNs obsolete and allowed the testing of a single DNN.

Ma et al. \cite{ma2018} extended the neuron coverage metric in two different ways to increase its granularity. First, they proposed k-multi-section coverage by partitioning the range of output values of each neuron at the training stage into equal sections. The k-multi-section neuron coverage metric then measures, how thoroughly neuron coverage is distributed over the sections. Second, for neuron boundary coverage, Ma et al. gauged how many regions beyond the defined output range have been covered. In addition, Ma et al. have demonstrated that their proposed coverage criteria scale well to practical-sized DNN models and are able to detect erroneous behavior triggered by state-of-the-art adversarial attack generation techniques.

Picking up modified condition / decision coverage (MC/DC)~\cite{hayhurst2001practical}, Sun et al. proposed a variant
in~\cite{sun2018} and further refined it in~\cite{sunHKSHA19}. For the test case generation, they used linear programming in~\cite{sun2018}, a gradient descent search algorithm in ~\cite{sunHKSHA19} and concolic testing in~\cite{sunWRHKK18}. Concolic testing is an alternative software testing that varies between concrete and symbolic execution. The core idea of MC/DC is that all possible conditions that contribute to a decision must be tested. Adapted to DNNs this means that not only the presence of a feature needs to be tested but also the effects of fewer complex features on a more complex feature must be tested.

%Cheng et al~\cite{chengHY18} suggested to partition the input domain of a DNN into equivalence classes and weight them based on their relative importance. Examples of such classes include They assume that these weighted class partitioning criteria already exist. For example, in the field of autonomous driving, these criteria can be based on weather, landscape, pedestrians, etc.
%Kim et al.~\cite{kimFY19} proposed a new metric independent from neuron coverage and this named their metric as Surprise Adequacy for DL systems. The metric measures how surprising the input is to a DL system with respect to the training data.
 
\subsection{Fuzzing for DNNs}

Fuzzing is a model examination with many inputs, which are generated according to a coverage metric and data mutation technique.

TensorFuzz, proposed by Odena et al.~\cite{odena2019}, was one of the first applications of coverage-guided fuzzing on neural networks. For the sampling of test data, a random selection and a sophisticated heuristic were used. The latter was motivated by the idea, that recently sampled and already mutated inputs that have been added back to the test suite are more likely to achieve higher coverage. The proposed mutator only applied additive white noise to the input images. Moreover, no coverage metrics were used, but neuron activations associated with a test input were stored and checked using an approximate nearest neighbors algorithm, to determine, whether other individuals or sets of already stored neuron activations are within a predefined distance.

DeepHunter by Xie et al.~\cite{xiema2019} is a coverage-guided grey-box DNN fuzzer, which is similar to TensorFuzz. Its major components are metamorphic mutation, DNN feedback, and batch pool maintenance. The goal of the proposed mutation strategy is to ensure that the semantics of the mutated and original images are the same for the viewer and that the resulting images are diversified and plausible. Coverage metrics used by Xie et al. include a classical neuron coverage metric and metrics from DeepGauge~\cite{ma2018}. To demonstrate the scaling ability of DeepHunter, the fuzzing technique was applied to practical-sized data sets like ImageNet and DNN models. %Eight applied mutations come from two categories, pixel value transformations (contrast, brightness, blur, noise) and affine transformations (translation, scaling, shearing, rotation). 

Pei et al.~\cite{pei2017} approach for detecting misbehavior in DNNs requires multiple DNNs for cross-referencing, resulting in an inability to scale to state-of-the-art networks. Guo et al.~\cite{guoJZCS18} proposed
a similar approach that requires a single DNN. The presented mutation algorithm applies a tiny perturbation to a test image, which is visibly indistinguishable. If the original and mutated test images are classified into different class labels, the prediction of the DNN is treated as incorrect behavior. The mutation algorithm is completed by solving a joint optimization problem of both maximizing neuron coverage and the number of incorrect behaviors. Furthermore, Guo et al. use several strategies for selecting neurons that are more likely to improve coverage.

\subsection{Criticism}%Recently a number of further coverage criteria has been introduced~\cite{tian2018}. Nevertheless,

Despite the initial success of coverage-based testing applied to DNNs, doubts also arose about the permeability of neural networks and the effectiveness of a true robustness improvement. Li et al.~\cite{li2019} criticized the use of coverage metrics for the generation of adversarial examples, as these do not offer any advantage or information gain compared to conventional methods such as FSGM. Li et al. further argue that these structural coverage criteria may be too coarse for adversarial inputs and at the same time too fine for misclassified natural inputs. The reasons include specific distribution of the adversarial examples in the data manifold and also the fact that the misclassified natural inputs are rare. Moreover, the experiments by Li et al. with natural inputs showed no correlation between the number of misclassified inputs in a test set and its structural coverage on the corresponding neural networks. %These results suggest that such criteria may be ineffective for error detection of DNNs with natural inputs. Their criticism can be summed up by saying that existing coverage metrics fail to capture and map the relationship between the network structure and the distributions of adversarial examples and misclassified natural inputs.

The negative criticism of the coverage metrics was taken up by Dong et al.~\cite{dong2019} in a large study with 100 DNNs for an image classification task using MNIST and  CIFAR datasets. Dong et al. examined the relationship between coverage and robustness for DNNs. However, they only considered intentional attacks with FGSM, JSMA, and C\&W. Various coverage metrics~\cite{pei2017, ma2018, kimFY19} were applied. In order to measure the robustness of the models, Dong et al. use the global Lipschitz constant and the CLEVER score. The correlation analysis concluded that if a model achieves high coverage of a metric accordingly, it is not necessarily robust and vice versa. Nor could it be claimed that retrained models, which then have higher coverage, have necessarily become more robust. Nevertheless, Dong et al.could show that the different coverage metrics do correlate with each other.

Two recent concurrent works \cite{harel2020} and \cite{yan2020} also demonstrated that increased coverage does not necessarily lead to better model quality.  Harel-Canada et al., in particular, demonstrated in \cite{harel2020}, that taking neuron coverage into consideration actually led to less test inputs found. The evaluation in the latter works is  restricted to MNIST and CIFAR datasets. In contrast, our work focuses on a more real-life use case of object detection.

\begin{figure*}[t]
\centering
	\includegraphics[width=0.8\textwidth]{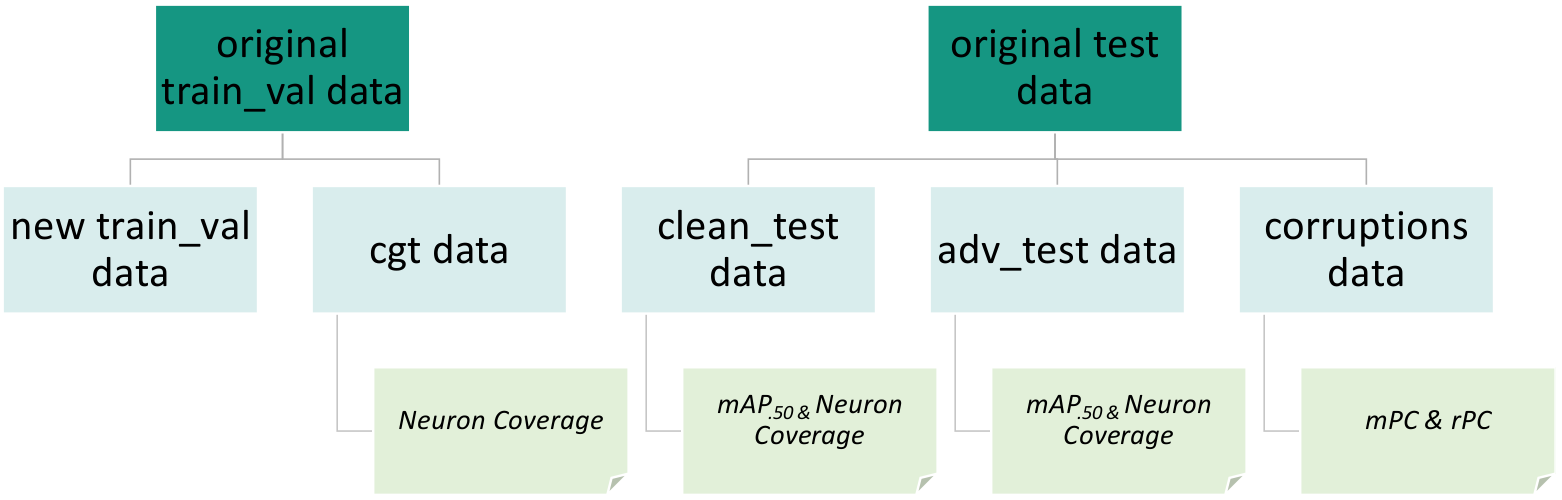}
	\caption{Dataset split with the corresponding evaluation metrics}
	\label{fig:dataset_split}
\end{figure*}

\section{CGT CONCEPT}

The proposed CGT pipeline consists of the following stages: (1) sampling random data for CGT, (2) mutating sampled data to generate bugs, (3) constructing a new training dataset and retraining the DNN. In the following, we describe all stages.

\subsection{Dataset Split}

To ensure fair evaluation of the retrained models and clear separation of CGT and training data, we define new subsets for training, validation, and testing (see Figure \ref{fig:dataset_split}).

For a given dataset with \texttt{train\_val} and \texttt{test} splits, we define new subsets as following. First, we randomly sample $\sfrac{2}{3}$ of the original \texttt{train\_val} -- this subset (\texttt{new train\_val}) will be further used to train and validate baselines. The remaining $\sfrac{1}{3}$ is reserved for CGT and is abbreviated as \texttt{cgt data} set.

For the original \texttt{test} data set partition, a copy of the data is maintained as \texttt{clean\_test data}. Then, an adversarial image is generated for each original test image resulting in a set called \texttt{adv\_test} data. Finally, \texttt{corruptions data} subset is created by applying corruptions proposed in Michaelis et al. \cite{michaelis2019}.

\subsection{Bug Definition}

According to Ma et al.~\cite{ma2018a}, DNN bugs can be divided into two categories: the first type is caused by suboptimal model structures, such as the number of hidden layers in a DNN model or the number of neurons in each layer, whereas the second type is caused by the misconducted training process (e.g. by using distorted training inputs). We further concentrate on the second type of DNN bug.

For the object detection use case, we apply the mean average precision (mAP) metric to measure DNN performance on clean ($x_{orig}$) and mutated ($x_{mut}$) inputs. We furthermore introduce a manually set parameter $\alpha_{mAP}$ to control bug severity. A set of bugs is then defined as follows:
\begin{equation}
bugs = \{x | x \in \texttt{cgt\_data}, {\frac{mAP(x_{mut})}{mAP(x_{orig})}} \leq \alpha_{mAP}\}
\label{eq:first-bug-def}
\end{equation}

Tian et al. were able to demonstrate empirically in their work that changes in
neuron coverage correlate with changes in the state of DNNs~\cite{tian2018}, so coverage metrics
can be used as a guiding mechanism for the systematic investigation of DNN states and behavior.
We further extend the bug definition to incorporate the condition that the coverage metric achieved by the mutated input should be higher compared to the coverage achieved by the original input:
\begin{equation}
\begin{split}
bugs = \{x | x \in \texttt{cgt\_data}, {\frac{mAP(x_{mut})}{mAP(x_{orig})}} \leq \alpha_{mAP}, \\
Cov({x_{orig}}) < Cov(x_{mut})\},
\end{split}
\label{eq:second-bug-def}
\end{equation}
where $Cov(x)$ stands for neuron coverage achieved on input $x$.

We use three options for $Cov_x$: neuron coverage (NC)~\cite{pei2017}, neuron boundary coverage (NBC)~\cite{ma2018} and strong neuron activation coverage (SNAC)~\cite{ma2018}.

For a DNN input $x$ and a set of all neurons $N$ in a DNN, an activated neuron is the one for which activation $act(n,x)$ on a given input $x$ exceeds some predefined threshold $t$. NC is then defined as a ratio of activated neurons:
\begin{equation}
\begin{split}
   NC(x) = \frac{|ActivatedNeurons(x)|}{|N|},\  \text{where} \\
   ActivatedNeurons(x) = \{n | act(n, x) > t\}
\end{split}
\end{equation}

For NBC, the range of output values for each neuron $n \in N$ is monitored on the training set (\texttt{ newtrain\_val} in our case) to define its upper ($high_n$) and lower ($low_n$) boundary output values. NBS for input $x$ is then defined as a ratio of neurons whose output values belong to their upper and lower corner regions for $x$:
\begin{equation}
\begin{split}
    NBC(x) = \frac{|UpperN(x)| + |LowerN(x)|}{2*|N|},\   \text{where}\\
    UpperN(x) = \{n \in N| act(n, x) \in (high_n, +\infty)\}\\
    LowerN(x) = \{n \in N| act(n, x) \in (-\infty, low_n)\}
\end{split}
\end{equation}

SNAC measures how many corner case regions above the major function region have been covered by a single test image:
\begin{equation}
    SNAC(x) = \frac{|UpperN(x)|}{N}
\end{equation}

\subsection{Pipeline Overview}
CGT starts with a baseline model, trained on \texttt{new train\_val}. Baseline performance is evaluated and stored for reference. In addition, neuron coverage metrics are computed on \texttt{clean\_test data} and  \texttt{adv\_test data}. To complete the initial evaluation, corruptions metrics \texttt{mPC} and \texttt{rPC} as defined in \cite{michaelis2019} are measured on \texttt{corruptions data}.

Next, images are randomly sampled from \texttt{cgt data} set. Natural image transformations are used for the simulation of unintended malfunctioning whereby adversarial examples are used for the simulation of intended malfunctioning. The detailed process of performing natural image transformations is described below. The sampled and mutated test inputs are at this point fully qualified to be used for coverage-guided testing.

To study intended misbehavior of a network, we attack it with adversarially mutated images during the testing phase. The adversarial images are created in advance to use them in combination with CGT. This is done to reduce the time required for the test process itself. There is no need for the second stage of testing for adversarial images, as enough errors are already exposed at the first stage.

For both test techniques each original clean image from the \texttt{cgt data} set is first passed to the network. Afterward, the mutated image is passed in order to calculate the relative performance change, and also the neuron coverage for both inputs. Subsequently it could be determined whether an image triggered misbehavior as defined in Equation \ref{eq:first-bug-def} or in Equation \ref{eq:second-bug-def} if neuron coverage is used. If misbehavior is observed, the input is added to the set of already encountered bugs.

Once all rounds are passed through, the retraining data is constructed with found bugs and a new \texttt{train\_val} data set. When retraining the model, it is crucial to use the same training procedure along with the hyperparameters as used for the baseline. In this way, it can be ruled out that possible performance increases are achieved by changing training parameters. 

In the last step, the performance is re-evaluated. If the results of the re-evaluation meet the performance and robustness requirements of the detector, the CGT process is terminated. %Otherwise the process can be repeated, but now with the retrained model as a baseline for re-testing.

\subsection{Generating Natural Transformations}

\begin{figure}[t]
\centering
	\includegraphics[width=0.3\textwidth]{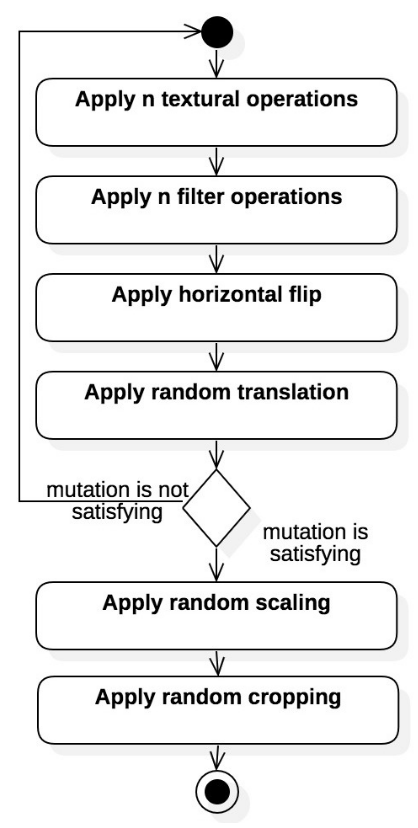}
	\caption{Overview of the mutation process}
	\label{fig:mutation}
\end{figure}

To robustify the model against unintended misbehavior, we incorporate a set of natural transformations in the mutation process. Figure \ref{fig:mutation} presents an overview of the mutation process for the CGT using natural image transformations. 

First, we apply a randomly selected subset of image operations from the set \textit{\{brightness, contrast, color, sharpness\}}. Each of these operations takes an enhancement factor parameter between 0.0 and 1.0. A parameter value of 1.0 returns the original image. A parameter value of 0.0 returns a black image in the case of the brightness operation, a grey image in the case of the contrast operation, a black and white image in the case of the color operation, and a blurred image in the case of the sharpness operation. %To allow a variety of different combinations of mutations, for each image the value of the enhancement factor parameter was randomly selected. Since DNN robustness against unseen inputs is to be investigated in the context of different, poorer image environments, it is advisable to select the interval close to 0.0.

After textural enhancement, a randomly selected subset of filter operations \textit{\{detail, edge\_enhance,  smooth, sharpen\}} is applied. An edge enhancement filter has the effect of increasing the contrast of pixels around specific edges so that after filter application edges are distinctively more visible. Smoothing filter operations are used to reduce noises present in the image and produce a less pixelated image. The sharpening filter emphasizes transitions between different regions in an image rather than being smooth. As an image passes through the sharpening filter, brighter pixels are amplified relative to adjacent pixels. %The filter operations form the second stage of textural operations and complete them. 

Next, horizontal flipping is applied. Retraining the DNN with horizontally flipped images can help to increase the invariance with respect to the orientation of an object \cite{Romera18}. Once the image is flipped, translation is applied. An image is randomly translated by 0-2 pixels since YOLOV3 employs $3\times3$ convolutions. Retraining the network with images that have been translated pixel by pixel and triggered a bug causes the network to view objects from different positions and thus does not always produce the same activations in the first layer \cite{Romera18}.

After textural and geometric transformations, an acceptance test for the mutations is applied. This is necessary to prevent the network from being tested with images that have lost their semantics due to the previous mutation process. At this point, there is the challenge to find a balance between mutated, but still recognizable images and a strong variety of mutations. Slightly mutated images also have a lower probability of triggering a bug. We apply the acceptance test proposed in \cite{xiema2019} and adapt it for our mutations sequence. 

If the mutation passed the acceptance test so that the previous mutation is metamorphic, the image is stretched and cropped in the very last step. Retraining with randomly scaled images that have triggered a bug, enables the model to see different scales of each object and improves network invariance with respect to different image resolutions. Random scaling is performed equally between 0.5 and 1.0 times the native resolution. Furthermore, the images are cropped, which has the same effect on the retrained model as translating the images. The now successful mutated image is saved and made available for the next step of CGT. 

In the case of an unsatisfactory mutation, i.e. a failed acceptance test, the image is run through the mutation sequence again. In the event of an unsuccessful mutation, further runs are started for an image. If still no successful mutation was found for that image, the image is discarded for this test round.

\subsection{CGT with Natural Transformations}

To discover malfunctions, the network is presented a pair of images: the original unmodified image and the mutated counterpart. The complete process of testing with unintentional attacks is divided into two stages. 

\textbf{The first stage} of the testing process is organized as follows. For each round $i$ of a run through the CGT process, $n$ images are sampled from \texttt{cgt data} set. Next, the sample is mutated and the corresponding adapted annotations are created. The testing loop starts by picking the head element of the list and predicting the bounding boxes for the original image. Once the prediction is completed, coverage and average precision are computed for the original test input. After completing the original input, both steps are repeated for the mutated image. The average precision ratio is then computed and the difference between both coverage values is obtained. If the ratio of average precision is less than or equal to $\alpha_{mAP}$ as well as the value of neuron coverage by the mutated image was greater than that of the original image, there is a bug according to Equation \ref{eq:second-bug-def} and the mutated and original images are both added to the set of all found bugs. Otherwise, CGT continues with the next image of the sample. After testing the network for in total $n\_rounds$ with $n\_samples$ per round $i$, the first stage of coverage-guided testing with naturally mutated inputs is completed. 

\textbf{The second stage} uses the entire data set resulting from the first stage as a test suite instead of sampling images from the \texttt{cgt data} set. In each round the complete test suite is mutated, i.e. no samples are taken from the test suite.

We assume, that errors found at the first phase are particularly useful and valuable for identifying network malfunctions. There is a possibility that images, which already caused an error at the first stage, could trigger another error after a new mutation at the second stage. Similar intuition was applied in \cite{odena2019}, where it was also assumed that inputs that have already been mutated once and have revealed misbehavior of the DNN are more likely to exhibit further different types of bugs of the DNN.

The acceptance test for mutations from the second stage is performed with the mutated image from the first stage as the original image to be referenced.

%\begin{figure}
%\centering
	%\includegraphics[width=0.3\textwidth]{images/sampler-mutator.png}
	%\caption{Sampler-mutator}
	%\label{fig:sampler-mutator}
%\end{figure}

%Final bugs in CGT with unintended attacks
%\begin{figure}
%	\includegraphics[width=0.5\textwidth]{images/unintended_bugs.png}
%	\caption{Sampler-mutator}
%	\label{fig:unintended-bugs}
%\end{figure}

\begin{figure*}
 \centering
	 \begin{subfigure}[b]{0.33\linewidth}
  	    \includegraphics[width=\textwidth]{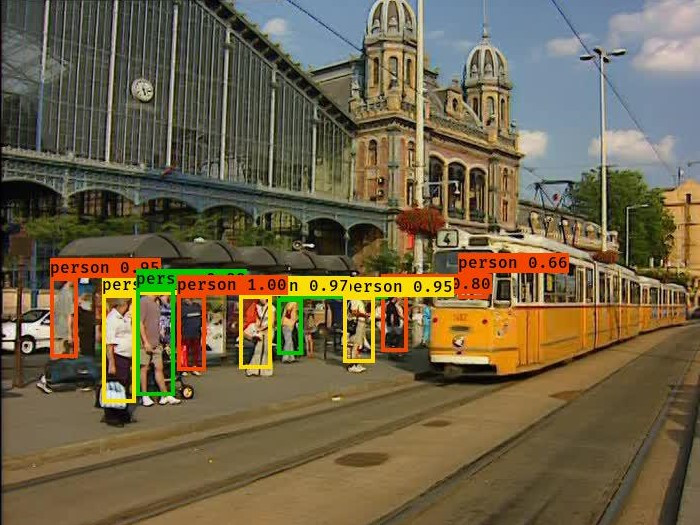}
  	    \caption{Clean input}
    \end{subfigure}
    \begin{subfigure}[b]{0.33\linewidth}
  	    \includegraphics[width=\textwidth]{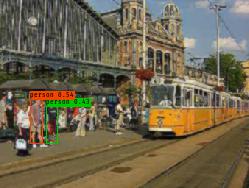}
  	    \caption{Mutated image at stage 1}
    \end{subfigure}
    \begin{subfigure}[b]{0.33\linewidth}
  	    \includegraphics[width=\textwidth]{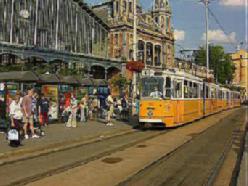}
  	    \caption{Mutated image at stage 2}
    \end{subfigure}
	\caption{Baseline predictions for the clean input and the corresponding bugs found via CGT with natural mutations}
	\label{fig:stages}
\end{figure*}

\section{EXPERIMENTS}
In the following, we present the evaluation of the proposed CGT pipeline on the person detection use case.

\subsection{Baseline and Models}
Experiments are performed on the \textit{CrowdHuman}~\cite{shao2018} dataset with 15K train/val and 4370 test images. This dataset focuses on crowded scenes in indoor and outdoor environments. The dataset splitting is performed as described above. Train/val data is split to 10K \texttt{new train\_val data} and 5K \texttt{cgt data} images. For the test subset, the further test datasets \texttt{clean\_test data}, \texttt{adv\_test data} and \texttt{corruptions data} of the same size are created. Performance is measured with the standard $mAP_{50}$ metric. The mAP measured on the \texttt{clean\_test data} is denoted $mAP_{clean}$, mAP measured on the \texttt{adv\_test data} -- $mAP_{adv}$.

For the baseline model, we used the pretrained \textit{Darknet53} weights, the evaluation was on the \texttt{new train\_val data}. The retrained model is trained on \texttt{new train\_val} with bugs added. The baseline reaches 43.45\% $mAP_{clean}$ and 28.71\% $mAP_{adv}$. On \texttt{corruptions data} the baseline reached 25.32\% $mPC$ and 58.41\% $rPC$. Note that we further report only relative improvement in mAP with respect to the indicated baseline performance.

Neuron coverage was evaluated with $t \in \{0.25, 0.5, 0.75\}$ and $t_{single}=0.5$ at \texttt{clean\_test data}, \texttt{adv\_test data} and \texttt{cgt data}. Table \ref{baseline-nc} shows the results for the baseline. The highest neuron coverage was achieved on the set of adversarially mutated images.

For natural mutations, we train \texttt{nm-} models as follows: three rounds with 1000 sampled images each at stage 1 and five rounds with a set of bugs from the first stage at stage 2. Figure  \ref{fig:stages} shows examples of bugs generated at different stages and baseline predictions for them.

For adversarial mutations, we train the \texttt{am-} models with three rounds of 500 sampled adversarial images each.  We use the \texttt{Daedalus} attack \cite{wang2019}, which is in turn based on the C\&W attack~\cite{carlini017}. The \texttt{Daedalus} attack aims to achieve an extremely dense false positives rate by triggering a malfunctioning of the Non-Maximum Suppression (NMS) component in an object detection system.

Overall, for each mutation technique (natural and adversarial), we obtain 8 retrained models with coverage metric and 2 models without coverage metric.

\begin{table}
     \begin{center}
 		\begin{tabular}{|r | c | c | c | }
 			\hline
 			\textbf{Dataset} & $t=.25$ & $t=.5$ & $t=.75$ \\
 			\hline \hline
 		    \texttt{clean\_test data} & 64.81 & 56.44 & 2.55 \\
 			\texttt{adv\_test data} & \textbf{69.27} & \textbf{57.04} & \textbf{3.38} \\
 			\texttt{corruptons data} & 63.66 & 56.10 & 2.49 \\
 			\hline
 		\end{tabular}
 \end{center}
\caption{Neuron coverage $NC$ achieved by the baseline}
\label{baseline-nc}
\end{table}

%\begin{figure}
%	 \begin{subfigure}[b]{0.4\linewidth}
%  	    \includegraphics[width=\textwidth]{images/273275,3f10a000a6668d95_r1_i143_org.jpg}
%    \end{subfigure}
%    \begin{subfigure}[b]{0.4\linewidth}
%  	    \includegraphics[width=\textwidth]{images/273275,3f10a000a6668d95_r1_i143_am.jpg}
%    \end{subfigure}%
	%\caption{Example of a bug found via adversarial attacks. From left to right: original image, mutation of the original image}
	%\label{fig:unintended-img}
%\end{figure}
%\begin{figure}
% \centering
%	 \begin{subfigure}[b]{0.48\linewidth}
%  	    \includegraphics[width=\textwidth]{images/273275,3f10a000a6668d95_r1_i143_org.jpg}
%    \end{subfigure}
%    \begin{subfigure}[b]{0.48\linewidth}
%  	    \includegraphics[width=\textwidth]{images/273275,3f10a000a6668d95_r1_i143_am.jpg}
%    \end{subfigure}%
	%\caption{Baseline predictions for the clean input (left) and for a corresponding bug found via CGT with adversarial mutations (right)}
	%\label{fig:unintended-img}
%\end{figure}

\begin{table}[t]
     \begin{center}
 		\begin{tabular}{|r | c |  c   c | c   c|}
 			\hline
 			
 			\textbf{Model /} & \textbf{$\alpha_{mAP}$} & \multicolumn{2}{|c}{\texttt{nm-}models} & \multicolumn{2}{|c|}{\texttt{am-}models} \\
 			 \textbf{Metric} &   & clean & adv & clean & adv \\
 			
 			\hline \hline
 			None & 0.6 & 25.81 & 58.34 & 26.94 & \textbf{68.41} \\
 			None & 0.3 & 26.04 & \textbf{68.58} & 27.38 & 65.97 \\
 			NC & 0.6 & 26.37 & 50.89  & 26.76 &61.93\\
 			NC & 0.3 & \textbf{26.41} & 54.89 & 24.38 & 64.23\\
 			SNA & 0.6 & 26.14 & 63.67 & 26.07 & 62.56 \\
 			SNA & 0.3 & 26.25 & 58.79 & 26.74 & 64.68\\
 			NB & 0.6 & 26.25 & 60.68 & \textbf{27.57} & 65.97\\
 			NB & 0.3 & \textbf{26.41} & 66.00 & 26.30 & 63.29\\
 			NB+SNA & 0.6 & 26.18 & 65.97  & 26.46 & 62.14\\
 			NB+SNA & 0.3 & 26.39 & 61.09 & 26.51 & 63.18\\
 			\hline
 		\end{tabular}
 \end{center}
\caption{Relative change in \textbf{mAP} performance on \textbf{clean} and \textbf{adversarial} data for models retrained with naturally (\texttt{nm-}) and adversarially (\texttt{am-}) mutated images compared to the baseline}
\label{map-nm-models}
\end{table}

%\begin{table}[t]
%     \begin{center}
% 		\begin{tabular}{|r | c | c | c | }
% 			\hline
% 			\textbf{Coverage Metric} & \textbf{$\alpha_{mAP}$} & \textbf{$mAP_{clean}$} & \textbf{$mAP_{adv}$} \\
% 			\hline \hline
% 			None & 0.3 & 26.94 & \textbf{68.41} \\
% 			None & 0.6 & 27.38 & \textbf{65.97} \\
% 			NC & 0.3 & 26.76 &61.93 \\
% 			NC & 0.6 & 24.38 & 64.23 \\
% 			SNA & 0.3 & 26.07 & 62.56 \\
% 			SNA & 0.6 & 26.74 & 64.68 \\
% 			NB & 0.3  & \textbf{27.57} & 65.97\\
% 			NB & 0.6 & 26.30 & 63.29 \\
% 			NB+SNA & 0.3 & 26.46 & 62.14 \\
% 			NB+SNA & 0.6 & 26.51 & 63.18 \\
% 			\hline
% 		\end{tabular}
% \end{center}
%\caption{Relative change in \textbf{mAP performance} of models retrained with \textbf{adversarially mutated} images compared to baseline.}
%\label{map-am-models}
%\end{table}

\begin{table}[t]
     \begin{center}
 		\begin{tabular}{|r | c | c  c | c c| }
 			\hline
 			\textbf{Model /} & \textbf{$\alpha_{mAP}$} & \multicolumn{2}{|c}{\texttt{nm-}models} & \multicolumn{2}{|c|}{\texttt{am-models}} \\
 			\textbf{Metric} & \textbf{$\alpha_{mAP}$} & \textbf{$mPC$} & \textbf{$rPC$} & \textbf{$mPC$} & \textbf{$rPC$}\\
 			\hline \hline
 			None & 0.6 & \textbf{41.07} & \textbf{11.92} & 35.39 & 6.28\\
 			None & 0.3 & 37.60 & 9.76 & 36.61 & 7.24\\
 			NC & 0.6 & 32.46 & 4.81 & 36.18 & 7.41\\
 			NC & 0.3 & 33.49 & 5.60 & 35.07 & \textbf{8.58} \\
 			SNA & 0.6 & 39.22 & 10.37  & 35.74 & 7.34\\
 			SNA & 0.3 & 35.82 & 7.69 & 36.65 & 7.82\\
 			NB & 0.6 & 37.09 & 8.58 & \textbf{37.24} & 7.57\\
 			NB & 0.3 & 38.74 & 9.72 & 35.27 & 7.10\\
 			NB+SNA & 0.6 & 39.45 & 10.51 & 36.53 & 8.03\\
 			NB+SNA & 0.3 & 37.40 & 8.63 & 34.76 & 6.52\\
 			\hline
 		\end{tabular}
 \end{center}
\caption{Relative change in \textbf{mPC and rPC performance} for models retrained with naturally (\texttt{nm-}) and adversarially (\texttt{am-}) mutated images compared to the baseline}
\label{mpc-nm-models}
\end{table}

%\begin{table}[t]
%     \begin{center}
% 		\begin{tabular}{|r | c | c | c | }
% 			\hline
% 			\textbf{Coverage Metric} & \textbf{$\alpha_{mAP}$} & \textbf{$mPC$} & \textbf{$rPC$} \\
% 			\hline \hline
% 			None & 0.3 & 35.39 & 6.28 \\
% 			None & 0.6 & 36.61 & 7.24 \\
% 			NC & 0.3 & 36.18 & 7.41 \\
% 			NC & 0.6 & 35.07 & 8.58 \\
% 			SNA & 0.3 & 35.74 & 7.34 \\
% 			SNA & 0.6 & 36.65 & 7.82 \\
% 			NB & 0.3 & \textbf{37.24} & 7.57 \\
% 			NB & 0.6 & 35.27 & 7.10 \\
% 			NB+SNA & 0.3 & 36.53 & 8.03 \\
% 			NB+SNA & 0.6 & 34.76 & 6.52 \\
% 			\hline
% 		\end{tabular}
% \end{center}
%\caption{Relative change in \textbf{mPC and rPC performance} of models retrained with \textbf{adversarially mutated} %images compared to baseline.}
%\label{mpc-am-models}
%\end{table}

\subsection{Robustness Improvement Results}

\begin{figure*}
 \centering
 \begin{subfigure}[t]{0.33\linewidth}
  	    \includegraphics[width=\textwidth]{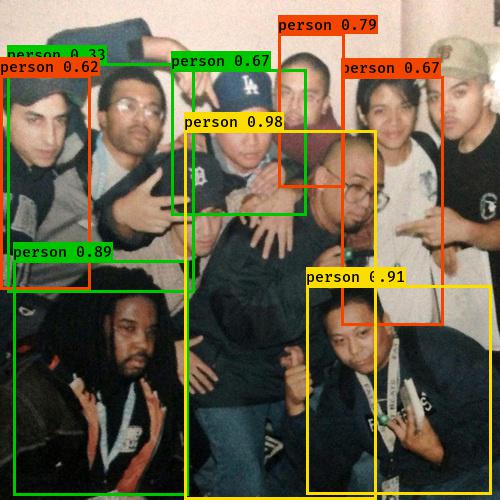}
  	    \caption{Baseline prediction for the clean input}
    \end{subfigure}
    %\begin{subfigure}[t]{0.24\linewidth}
  	%    \includegraphics[width=\textwidth]{images/nm_stage1.jpg}
  	%    \caption{Baseline prediction for the mutation of the original image at stage 1}
    %\end{subfigure}
    \begin{subfigure}[t]{0.33\linewidth}
  	    \includegraphics[width=\textwidth]{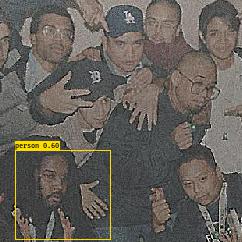}
  	    \caption{Baseline prediction for the mutation of the original image at stage 2}
    \end{subfigure}
    \begin{subfigure}[t]{0.33\linewidth}
  	    \includegraphics[width=\textwidth]{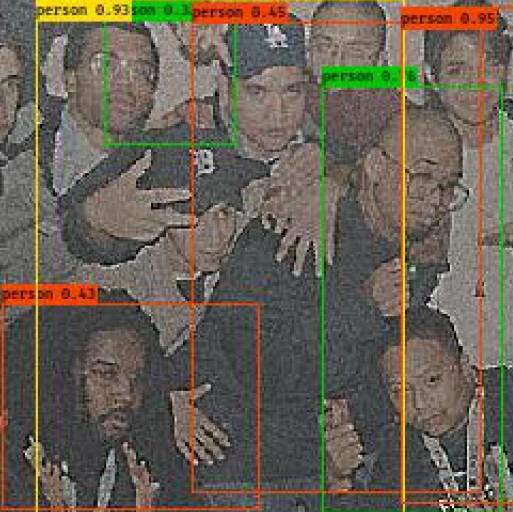}
  	    \caption{Prediction of the retrained model}
    \end{subfigure}
	\caption{Predictions for the clean input and a bug found via CGT using natural mutations}
	\label{fig:unintended-img}
\end{figure*}

\begin{figure*}
 \centering
  \begin{subfigure}[t]{0.33\linewidth}
  	    \includegraphics[width=\textwidth]{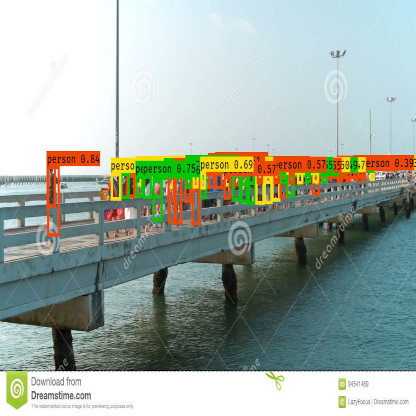}
  	    \caption{Baseline prediction for the clean input}
    \end{subfigure}
     \begin{subfigure}[t]{0.33\linewidth}
  	    \includegraphics[width=\textwidth]{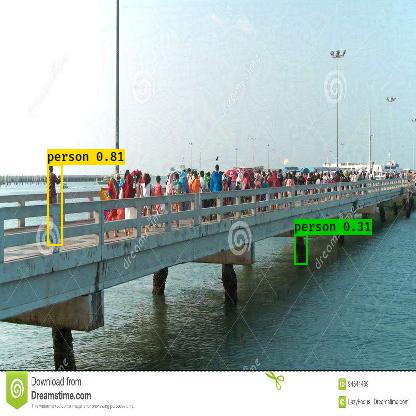}
  	    \caption{Baseline prediction for the adversarially mutated input}
    \end{subfigure}
     \begin{subfigure}[t]{0.33\linewidth}
  	    \includegraphics[width=\textwidth]{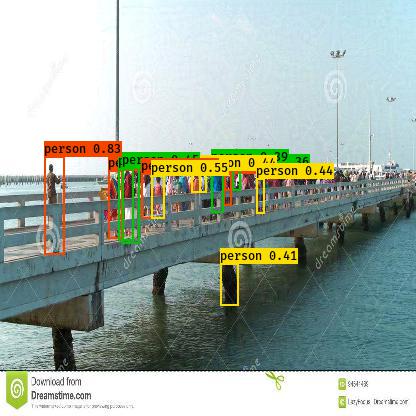}
  	    \caption{Prediction of the retrained model}
    \end{subfigure}
	\caption{Predictions for the clean input and a bug found via CGT using adversarial mutations}
	\label{fig:intended-img}
\end{figure*}

Table \ref{map-nm-models} demonstrates relative change in $mAP$ performance for the \texttt{nm-} and \texttt{am-} models. Unexpectedly, the model retrained with natural bugs achieves a significant improvement in robustness against adversarial images. Moreover, higher $\alpha_{mAP}$ led to better results, meaning that it is more beneficial to include in retraining more bugs, which are less severe. Also, no significant difference is gained when the coverage metric is excluded. As for the \texttt{am-} models, they exhibit greater enhancement of the robustness against adversarial attacks, as expected. Overall, CGT has managed to improve DNN robustness against natural and adversarial perturbations. Figures \ref{fig:unintended-img} and \ref{fig:intended-img}  further illustrate the enhanced performance of the retrained models.

We further report results on robustness against corrupted inputs in Table \ref{mpc-nm-models}. Apparently retraining DNNs with naturally mutated images helps to enhance this type of robustness. Also, the mean relative change of $mPC$ performance for \texttt{nm-} models is greater than the corresponding mean relative change of $mAP$ performance on clean data of the same models, indicating stronger robustness against corruptions of retrained models. For \texttt{am-} models, the improvement in robustness against corruptions is minimally lower compared to mean values of \texttt{nm-}models. Nevertheless, the increase remains remarkable considering that the bugs that were included in their retraining only contained adversarially perturbed images. Again, there is no significant difference by excluding the coverage metric.

\subsection{Impact of Neuron Coverage}

Next, we evaluate the impact of neuron coverage on increasing the robustness. Table \ref{map-relative} shows mean relative $mAP$ changes for models with and without coverage metric involved in testing. No significant gain could be achieved via the usage of coverage metrics. The analysis regarding robustness against corruptions (see Table \ref{mpc-relative}) is also consistent with these findings.

Furthermore, we have assessed the neuron coverage of the retrained models by comparing it to that of the baseline. We have observed, that the greatest relative deviations (positive and negative) from baseline occur for accumulated coverage with a threshold value of $t=.75$. Overall, most models have demonstrated the drop of the neuron coverage. However, this decrease was not consistent with the measured model robustness -- some models have still demonstrated higher neuron coverage compared to the baseline. These results are consistent with the work by Dong et al.~\cite{dong2019}, which also demonstrates that increased network robustness does not necessarily correlate with lower neuron coverage.

%\begin{table}[t]
%     \begin{center}
% 		\begin{tabular}{|r | c  c  c | c c c| }
% 			\hline
% 			\textbf{Model /} & \multicolumn{3}{|c}{\texttt{nm-}models} & \multicolumn{3}{|c|}{\texttt{am-models}} \\
% 			\textbf{Metric} & $t=.25$ & $t=.5$ & $t=.75$ & $t=.25$ & $t=.5$ & $t=.75$ \\
% 			\hline \hline
% 			1 & 0.12 & 0.01 & \textbf{11.92} & 35.39 & 6.28\\
% 			2 & 0.73 & -0.03 & 9.76 & 36.61 & 7.24\\
% 			3 & 1.77 & 0.00 & 4.81 & 36.18 & 7.41\\
% 			4 & 1.17 & 0.04 & 5.60 & 35.07 & \textbf{8.58} \\
% 			5 & 2.68 & 0.00 & 10.37  & 35.74 & 7.34\\
% 			6 & -0.29 & -0.03 & 7.69 & 36.65 & 7.82\\
% 			7 & 0.41 & 37.09 & 8.58 & \textbf{37.24} & 7.57\\
% 			8 & 0.58 & 38.74 & 9.72 & 35.27 & 7.10\\
% 			9 & 0.29 & 39.45 & 10.51 & 36.53 & 8.03\\
% 			10 & 0.58 & 37.40 & 8.63 & 34.76 & 6.52\\
% 			\hline
% 		\end{tabular}
% \end{center}
%\caption{Relative change in \textbf{neuron coverage} for models retrained with naturally (\texttt{nm-}) and adversarially (\texttt{am-}) mutated images compared to the baseline}
%\label{mpc-nm-models}
%\end{table}

Finally, Table \ref{all-relative} depicts the mean improvements of the different types of robustness achieved by applying the presented framework without distinction on the coverage metric usage. The best results were achieved for the adversarial attacks and models trained with adversarially mutated data. Overall, all three types of robustness have improved concurrently, the differences due to the nature of bugs involved in the training were not significant.

\begin{table}[t]
     \begin{center}
 		\begin{tabular}{|r | c | c | c | c|}
 			\hline
 			\textbf{Models} & \multicolumn{2}{c|}{$mAP_{clean}$} & \multicolumn{2}{c|}{$mAP_{adv}$} \\
 			 & cov & no cov & cov & no cov \\
 			\hline \hline
 		    \texttt{nm-} & 26.29 & 25.93 & 60.25 & 63.46 \\
 			\texttt{am-} & 26.35 & \textbf{27.16} & 63.50 & \textbf{67.19} \\
 			\hline
 		\end{tabular}
 \end{center}
\caption{Mean relative change in \textbf{mAP performance} for models retrained with and without
coverage metrics on \texttt{clean\_test data} and \texttt{adv\_test data}}
\label{map-relative}
\end{table}

\begin{table}[t]
     \begin{center}
 		\begin{tabular}{|r | c | c | c | c|}
 			\hline
 			\textbf{Models } &  \multicolumn{2}{c|}{$mPC$} & \multicolumn{2}{c|}{$rPC$} \\
 			 & with cov & no cov & with cov & no cov \\
 			\hline \hline
 		    \texttt{nm-} & 36.71 & \textbf{39.34} & 8.24 & \textbf{10.84} \\
 			\texttt{am-} & 35.93 & 36.38 & 7.55 & 7.24 \\
 			\hline
 		\end{tabular}
 \end{center}
\caption{Mean relative change in \textbf{mPC and rPC performance} for models retrained with and without
coverage metrics}
\label{mpc-relative}
\end{table}

\begin{table}[t]
     \begin{center}
 		\begin{tabular}{|r | c | c |}
 			\hline
 			\textbf{Robustness against ... } &  \texttt{nm-}models & \texttt{am-}models \\
 			\hline \hline
 		    naturally mutated data & 26.21 & 26.51 \\
 			corrupted data & 37.24 & 35.90 \\
 			adversarial data & 60.89 & 64.24 \\
 			\hline
 		\end{tabular}
 \end{center}
\caption{Mean relative improvements in \textbf{mAP} (in \%) averaged over all \texttt{nm-} and all \texttt{am-} models}
\label{all-relative}
\end{table}


\section{CONCLUSION}

In this work, we have applied coverage-guided testing (CGT) to the task of person detection using deep neural networks. For this, we have proposed a method to find bugs by introducing natural and adversarial mutations to the inputs, randomly sampled from a dedicated subset of the training data. The discovered bugs were then included in the DNN retraining process to enhance the robustness of the analyzed models. We have considered two types of bugs: those which lead to decreased performance and those which additionally lead to increased coverage. The particular focus of the work was on the necessity of including coverage metrics in the bug definition. For this, three popular neuron coverage metrics as well as a combination of them were considered.

We have evaluated the proposed fuzzer on the YOLOv3 model, trained on the CrowdHuman dataset. We could successfully find numerous bugs both via natural and adversarial perturbations. Models retrained on a dataset with bugs have demonstrated enhanced robustness to naturally mutated, corrupted, and adversarial inputs. Interestingly, the robustness to several misbehavior types could be reached, even though bugs used for retraining covered only one of them. However, we could not prove the clear adequacy of the coverage metrics. Our experiments have shown that robustness improvements can be achieved without the inclusion of a coverage metric into the bug definition. This casts doubt on the effectiveness of the coverage metrics in CGT for object detection and is consistent with recent findings for other computer vision tasks.

%\addtolength{\textheight}{-2cm}   % This command serves to balance the column lengths
                                  % on the last page of the document manually. It shortens
                                  % the textheight of the last page by a suitable amount.
                                  % This command does not take effect until the next page
                                  % so it should come on the page before the last. Make
                                  % sure that you do not shorten the textheight too much.

\section*{Acknowledgement}

The research leading to these results is funded by the German Federal Ministry for Economic Affairs and Climate Action" within the project “Methoden und Maßnahmen zur Absicherung von KI basierten Wahrnehmungsfunktionen für das automatisierte Fahren (KI-Absicherung)". The authors would like to thank the consortium for the successful cooperation.

%%%%%%%%% REFERENCES
{\small
\bibliographystyle{ieee_fullname}
\bibliography{main}
}

\end{document}